\documentclass[11pt,a4paper]{article}
\usepackage[latin9]{inputenc}
\usepackage{amsmath}
\usepackage{graphicx}
\usepackage[authoryear]{natbib}
\PassOptionsToPackage{normalem}{ulem}
\usepackage{ulem}
\usepackage[unicode=true]
 {hyperref}

\makeatletter

\pdfpageheight\paperheight
\pdfpagewidth\paperwidth

\newcommand{\noun}[1]{\textsc{#1}}
\providecommand{\tabularnewline}{\\}

\@ifundefined{date}{}{\date{}}
\usepackage[hyperref]{acl2017}
\usepackage{times}
\usepackage{latexsym}
\usepackage{amsmath}
\usepackage{amssymb}
\usepackage{url}
\usepackage[font={small}]{caption}

\aclfinalcopy % Uncomment this line for the all SemEval submissions
\renewcommand{\and}{\end{tabular}\kern-\tabcolsep\ and\ \kern-\tabcolsep\begin{tabular}[t]{c}}

\title{Amobee at SemEval-2017 Task 4:  Deep Learning System for Sentiment Detection on Twitter}
\author{Alon Rozental\thanks{~~These authors contributed equally to this work.}~~, Daniel Fleischer\footnotemark[1] \\
  Amobee, Tel Aviv, Israel \\
  {\tt \{arozental,danielf\}@amobee.com} }

\makeatother

\begin{document}
\maketitle 
\begin{abstract}
This paper describes the Amobee sentiment analysis system, adapted
to compete in SemEval 2017 task 4. The system consists of two parts:
a supervised training of RNN models based on a Twitter sentiment treebank,
and the use of feedforward NN, Naive Bayes and logistic regression
classifiers to produce predictions for the different sub-tasks. The
algorithm reached the 3rd place on the 5-label classification task
(sub-task C).
\end{abstract}

\section{Introduction}

Sentiment detection is the process of determining whether a text has
a positive or negative attitude toward a given entity (topic) or in
general. Detecting sentiment on Twitter\textemdash a social network
where users interact via short 140-character messages, exchanging
information and opinions\textemdash is becoming ubiquitous. Sentiment
in Twitter messages (tweets) can capture the popularity level of political
figures, ideas, brands, products and people. Tweets and other social
media texts are challenging to analyze as they are inherently different;
use of slang, mis-spelling, sarcasm, emojis and co-mentioning of other
messages pose unique difficulties. Combined with the vast amount of
Twitter data (mostly public), these make sentiment detection on Twitter
a focal point for data science research. 

SemEval is a yearly event in which teams compete in natural language
processing tasks. Task 4 is concerned with sentiment analysis in Twitter;
it contains five sub-tasks which include classification of tweets
according to 2, 3 or 5 labels and quantification of sentiment distribution
regarding topics mentioned in tweets; for a complete description of
task 4 see \citet{SemEval:2017:task4}. 

This paper describes our system and participation in all sub-tasks
of SemEval 2017 task 4. Our system consists of two parts: a recurrent
neural network trained on a private Twitter dataset, followed by a
task-specific combination of model stacking and logistic regression
classifiers.

The paper is organized as follows: section \ref{sec:RNN-Models} describes
the training of RNN models, data being used and model selection; section
\ref{sec:Features-Extraction} describes the extraction of semantic
features; section \ref{sec:Experiments} describes the task-specific
workflows and scores. We review and summarize in section \ref{sec:Review-and-Conclusions}.
Finally, section \ref{sec:Future-Work} describes our future plans,
mainly the development of an LSTM algorithm. 

\section{\label{sec:RNN-Models}RNN Models}

The first part of our system consisted of training recursive-neural-tensor-network
(RNTN) models \citep{socher2013recursive}. 

\subsection{Data}

Our training data for this part was created by taking a random sample\footnote{We used Twitter stream API.}
from Twitter and having it manually annotated on a 5-label basis to
produce fully sentiment-labeled parse-trees, much like the Stanford
sentiment treebank. The sample contains twenty thousand tweets with
sentiment distribution as following:\medskip{}

\begin{center}
\begin{tabular}{|c|c|c|c|c|c|}
\cline{2-6} 
\multicolumn{1}{c|}{} & {\small{}v-neg.} & {\small{}neg.} & {\small{}neu.} & {\small{}pos.} & {\small{}v-pos.}\tabularnewline
\hline 
{\small{}Train} & {\small{}$8.4\%$} & {\small{}$23.2\%$} & {\small{}$31.7\%$} & {\small{}$25.3\%$} & {\small{}$11.4\%$}\tabularnewline
\hline 
{\small{}Test} & {\small{}$8.6\%$} & {\small{}$23.0\%$} & {\small{}$33.2\%$} & {\small{}$24.8\%$} & {\small{}$10.4\%$}\tabularnewline
\hline 
\end{tabular}\medskip{}
\par\end{center}

\subsection{Preprocessing}

First we build a custom dictionary by means of crawling Wikipedia
and extracting lists of brands, celebrities, places and names. The
lists were then pruned manually. Then we define the following steps
when preprocessing tweets:
\begin{enumerate}
\item Standard tokenization of the sentences, using the Stanford coreNLP
tools \citep{manning-EtAl:2014:P14-5}.
\item Word-replacement step using the Wiki dictionary with representative
keywords.
\item Lemmatization, using coreNLP.
\item Emojis: removing duplicate emojis, clustering them according to sentiment
and replacing them with representative keywords, e.g. ``happy-emoji''. 
\item Regex: removing duplicate punctuation marks, replacing URLs with a
keyword, removing Camel casing. 
\item Parsing: parts-of-speech and constituency parsing using a shift-reduce
parser\footnote{\href{http://nlp.stanford.edu/software/srparser.shtml}{http://nlp.stanford.edu/software/srparser.shtml}.},
which was selected for its speed over accuracy. 
\item NER: using entity recognition annotator\footnote{\href{http://nlp.stanford.edu/software/CRF-NER.shtml}{http://nlp.stanford.edu/software/CRF-NER.shtml}.},
replacing numbers, dates and locations with representative keywords. 
\item Wiki: second step of word-replacement using our custom wiki dictionary.
\end{enumerate}

\subsection{Training}

We used the Stanford coreNLP sentiment annotator, introduced by \citet{socher2013recursive}.
Words are initialized either randomly as $d$ dimensional vectors,
or given externally as word vectors. We used four versions of the
training data; with and without lemmatization and with and without
pre-trained word representations\footnote{Twitter pre-trained word vectors were used, \href{http://nlp.stanford.edu/projects/glove/}{http://nlp.stanford.edu/projects/glove/}}
\citep{pennington2014glove}.

\subsection{Tweet Aggregation}

Twitter messages can be comprised of several sentences, with different
and sometimes contrary sentiments. However, the trained models predict
sentiment on individual sentences. We aggregated the sentiment for
each tweet by taking a linear combination of the individual sentences
comprising the tweet with weights having the following power dependency:
\begin{align}
h(f,l,\mbox{pol})=(1+f)^{\alpha}\,l^{\,\beta}\,(1+\mbox{pol})^{\gamma}+1,
\end{align}
where $\alpha,\beta,\gamma$ are numerical factors to be found, $f,l,\mbox{pol}$
are the fraction of known words, length of the sentence and polarity,
respectively, with polarity defined by: 
\begin{align}
\textrm{pol}=\left|10\cdot\text{vn}+\text{n}-\text{p}-10\cdot\text{vp}\right|,
\end{align}
where vn, n, p, vp are the probabilities as assigned by the RNTN for
very-negative, negative, positive and very-positive label for each
sentence. We then optimized the parameters $\alpha,\beta,\gamma$
with respect to the true labels.

\subsection{Model Selection}

After training dozens of models, we chose to combine only the best
ones using stacking, namely combining the models output using a supervised
learning algorithm. For this purpose, we used the Scikit-learn \citep{scikit-learn}
recursive feature elimination (RFE) algorithm to find both the optimal
number and the actual models, thus choosing the best \uline{five}
models. The models chosen include a representative from each type
of the data we used and they were:
\begin{itemize}
\item Training data without lemmatization step, with randomly initialized
word-vectors of size 27.
\item Training data with lemmatization step, with pre-trained word-vectors
of size 25.
\item 3 sets of training data with lemmatization step, with randomly initialized
word-vectors of sizes 24, 26.
\end{itemize}
The five models output is concatenated and used as input for the various
tasks, as described in \ref{subsec:General-Workflow}.

\section{\label{sec:Features-Extraction}Features Extraction}

In addition to the RNN trained models, our system includes feature
extraction step; we defined a set of lexical and semantical features
to be extracted from the original tweets: 
\begin{itemize}
\item In-subject, In-object: whether the entity of interest is in the subject
or object. 
\item Containing positive/negative adjectives that describe the entity of
interest. 
\item Containing negation, quotations or perfect progressive forms. 
\end{itemize}
For this purpose, we used the Stanford deterministic coreference resolution
system \citep{lee2011stanford,recasens_demarneffe_potts2013}.

\section{\label{sec:Experiments}Experiments}

The experiments were developed by using Scikit-learn machine learning
library and Keras deep learning library with TensorFlow backend \citep{abadi2016tensorflow}.
Results for all sub-tasks are summarized in table 
\begin{table*}
\begin{centering}
\emph{\noun{}}%
\begin{tabular}{|c|c|c|c|c|c|}
\hline 
{\small{}Task} & {\small{}A } & {\small{}B} & {\small{}C} & {\small{}D} & {\small{}E}\tabularnewline
 & {\small{}3-class.} & {\small{}2-class.} & {\small{}5-class.} & {\small{}2-quant.} & {\small{}5-quant.}\tabularnewline
\hline 
{\small{}Metric} & $\rho$ & $\rho$ & {\small{}$MAE^{M}$} & {\small{}$KLD$} & {\small{}$EMD$}\tabularnewline
\hline 
\hline 
{\small{}Score} & {\small{}$0.575$} & {\small{}$0.822$} & {\small{}$0.599$} & {\small{}$0.149$} & {\small{}$0.345$}\tabularnewline
\hline 
{\small{}Rank} & {\small{}27/37} & {\small{}11/23} & {\small{}3/15} & {\small{}11/15} & {\small{}6/12}\tabularnewline
\hline 
\end{tabular}
\par\end{centering}
\caption{\label{tab:Evaluation-results1}Summary of evaluation results, metrics
used and rank achieved, for all sub tasks. $\rho$ is macro-averaged
recall, $MAE^{M}$ is macro-averaged mean absolute error, $KLD$ is
Kullback-Leibler divergence and $EMD$ is earth-movers distance.}
\end{table*}
\ref{tab:Evaluation-results1}.

\subsection{General Workflow\label{subsec:General-Workflow}}

For each tweet, we first ran the RNN models and got a 5-category probability
distribution from each of the trained models, thus a 25-dimensional
vector. Then we extracted sentence features and concatenated them
with the RNN vector. We then trained a Feedforward NN which outputs
a 5-label probability distribution for each tweet. That was the starting
point for each of the tasks; we refer to this process as the pipeline.

\subsection{Task A}

The goal of this task is to classify tweets sentiment into three classes
(negative, neutral, positive) where the measured metric is a macro-averaged
recall. 

We used the SemEval 2017 task A data in the following way: using SemEval
2016 TEST as our TEST, partitioning the rest into TRAIN and DEV datasets.
The test dataset went through the previously mentioned pipeline, getting
a 5-label probability distribution. 

We anticipated the sentiment distribution of the test data would be
similar to the training data\textemdash as they may be drawn from
the same distribution. Therefore we used re-sampling of the training
dataset to obtain a skewed dataset such that a logistic regression
would predict similar sentiment distributions for both the train and
test datasets. Finally we trained a logistic regression on the new
dataset and used it on the task A test set. We obtained a macro-averaged
recall score of $\rho=0.575$ and accuracy of $Acc=0.587$. 

Apparently, our assumption about distribution similarity was misguided
as one can observe in the next table.

\medskip{}

\begin{center}
\begin{tabular}{|c|c|c|c|}
\cline{2-4} 
\multicolumn{1}{c|}{} & {\small{}Negative} & {\small{}Neutral} & {\small{}Positive}\tabularnewline
\hline 
{\small{}Train} & $15.5\%$ & $41.1\%$ & $43.4\%$\tabularnewline
\hline 
{\small{}Test} & $32.3\%$ & $48.3\%$ & $19.3\%$\tabularnewline
\hline 
\end{tabular}
\par\end{center}

\subsection{Tasks B, D}

The goals of these tasks are to classify tweets sentiment regarding
a given entity as either positive or negative (task B) and estimate
sentiment distribution for each entity (task D). The measured metrics
are macro-averaged recall and KLD, respectively. 

We started with the training data passing our pipeline. We calculated
the mean distribution for each entity on the training and testing
datasets. We trained a logistic regression from a 5-label to a binary
distribution and predicted a positive probability for each entity
in the test set. This was used as a prior distribution for each entity,
modeled as a Beta distribution. We then trained a logistic regression
where the input is a concatenation of the 5-labels with the positive
component of the probability distribution of the entity's sentiment
and the output is a binary prediction for each tweet. Then we chose
the label\textemdash using the mean positive probability as a threshold.
These predictions are submitted as task B. We obtained a macro-averaged
recall score of $\rho=0.822$ and accuracy of $Acc=0.802$. 

Next, we took the predictions mean for each entity as the likelihood,
modeled as a Binomial distribution, thus getting a Beta posterior
distribution for each entity. These were submitted as task D. We obtained
a score of $KLD=0.149$. 

\subsection{Tasks C, E}

The goals of these tasks are to classify tweets sentiment regarding
a given entity into five classes\textemdash very negative, negative,
neutral, positive, very positive\textemdash (task C) and estimate
sentiment distribution over five classes for each entity (task E).
The measured metrics are macro-averaged MAE and earth-movers-distance
(EMD), respectively. 

We first calculated the mean sentiment for each entity. We then used
bootstrapping to generate a sample for each entity. Then we trained
a logistic regression model which predicts a 5-label distribution
for each entity. We modified the initial 5-label probability distribution
for each tweet using the following formula: 
\begin{align}
p^{\text{new}}(t_{0},c_{0}) & =\sum_{c\in C}\frac{p\left(t_{0},c\right)\cdot p^{\text{entity-LR}}\left(t_{0},c_{0}\right)}{\sum_{t\in T}p\left(t,c\right)},
\end{align}
where $t_{0},c_{0}$ are the current tweet and label, $p^{\text{entity-LR}}$
is the sentiment prediction of the logistic regression model for an
entity, $T$ is the set of all tweets and $C=\left\{ \text{vn, n, neu, p, vp}\right\} $
is the set of labels. We trained a logistic regression on the new
distribution and the predictions were submitted as task C. We obtained
a macro-averaged MAE score of $MAE^{M}=0.599$.

Next, we defined a loss function as follows:
\begin{align}
\text{loss}(t_{0},c_{0}) & =\sum_{c\in C}\left|c-c_{0}\right|\cdot\frac{p\left(t_{0},c\right)}{\sum_{t\in T}p\left(t,c\right)},
\end{align}
where the probabilities are the predicted probabilities after the
previous logistic regression step. Finally we predicted a label for
each tweet according to the lowest loss, and calculated the mean sentiment
for each entity. These were submitted as task E. We obtained a score
of $EMD=0.345$. 

\section{\label{sec:Review-and-Conclusions}Review and Conclusions}

In this paper we described our system of sentiment analysis adapted
to participate in SemEval task 4. The highest ranking we reached was
third place on the 5-label classification task. Compared with classification
with 2 and 3 labels, in which we scored lower, and the fact we used
similar workflow for tasks A, B, C, we speculate that the relative
success is due to our sentiment treebank ranking on a 5-label basis.
This can also explain the relatively superior results in quantification
of 5 categories as opposed to quantification of 2 categories.

Overall, we have had some unique advantages and disadvantages in this
competition. On the one hand, we enjoyed an additional twenty thousand
tweets, where every node of the parse tree was labeled for its sentiment,
and also had the manpower to manually prune our dictionaries, as well
as the opportunity to get feedback from our clients. On the other
hand, we did not use any user information and/or metadata from Twitter,
nor did we use the SemEval data for training the RNTN models. In addition,
we did not ensemble our models with any commercially or freely available
pre-trained sentiment analysis packages.

\section{\label{sec:Future-Work}Future Work}

We have several plans to improve our algorithm and to use new data.
First, we plan to extract more semantic features such as verb and
adverb classes and use them in neural network models as additional
input. Verb classification was used to improve sentiment detection
\citep{chesley2006using}; we plan to label verbs according to whether
their sentiment changes as we change the tense, form and active/passive
voice. Adverbs were also used to determine sentiment \citep{benamara2007sentiment};
we plan to classify adverbs into sentiment families such as intensifiers
(``very''), diminishers (``slightly''), positive (``delightfully'')
and negative (``shamefully'').

Secondly, we can use additional data from Twitter regarding either
the users or the entities-of-interest. 
\begin{figure*}
\begin{centering}
\emph{\includegraphics[scale=0.35]{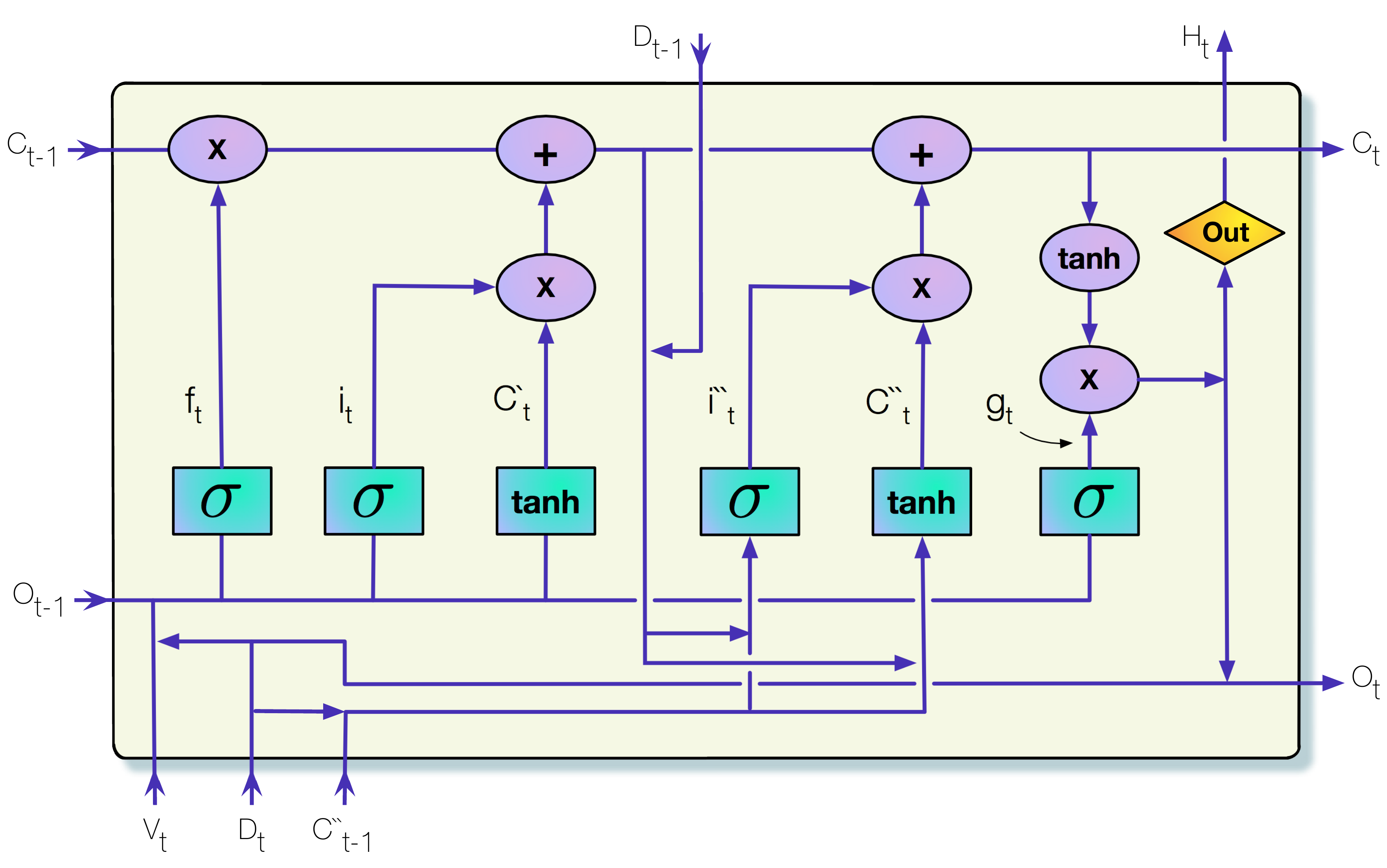}}
\par\end{centering}
{\small{}\caption{\label{fig:LSTM-module} LSTM module; round purple nodes are element-wise
operations, turquoise rectangles are neural network layers, orange
rhombus is a dim-reducing matrix, splitting line is duplication, merging
lines is concatenation.}
}{\small \par}
\end{figure*}

Finally, we plan to implement a long short-term memory (LSTM) network
\citep{hochreiter1997long} which trains on a sentence together with
all the syntax and semantic features extracted from it. There is some
work in the field of semantic modeling using LSTM, e.g. \citet{Palangi:2014aa,Palangi:2016:DSE:2992449.2992457}.
Our plan is to use an LSTM module to extend the RNTN model of \citet{socher2013recursive}
by adding the additional semantic data of each phrase and a reference
to the entity-of-interest. An illustration of the computational graph
for the proposed model is presented in figure \ref{fig:LSTM-module}.
The inputs/outputs are: $V$ is a word vector representation of dimension
$d$, $D$ encodes the parts-of-speech (POS) tagging, syntactic category
and an additional bit indicating whether the entity-of-interest is
present in the expression\textemdash all encoded in a $7$ dimensional
vector, $C$ is a control channel of dimension $d$, $O$ is an output
layer of dimension $d+7$ and $H$ is a sentiment vector of dimension
$s$. 

The module functions are defined as following:

\begin{align}
f_{t} & =\sigma\left[L_{f}\left(\left[V_{t},D_{t}\right],O_{t-1}\right)\right]\nonumber \\
i_{t} & =\sigma\left[L_{i}\left(\left[V_{t},D_{t}\right],O_{t-1}\right)\right]\nonumber \\
C'_{t} & =\tanh\left[L_{C'}\left(\left[V_{t},D_{t}\right],O_{t-1}\right)\right]\nonumber \\
i''_{t} & =\sigma\left[L_{i''}\left(\left[C''_{t-1},D_{t}\right],\left[C_{t-1},D_{t-1}\right]\right)\right]\nonumber \\
C''_{t} & =\tanh\left[L_{C''}\left(\left[C''_{t-1},D_{t}\right],\left[C_{t-1},D_{t-1}\right]\right)\right]\nonumber \\
g_{t} & =\sigma\left[L_{g}\left(\left[V_{t},D_{t}\right],O_{t-1}\right)\right]\nonumber \\
C_{t} & =C_{t-1}\odot f_{t}+C'_{t}\odot i_{t}+i''_{t}\odot C''_{t}\nonumber \\
H_{t} & =W_{\text{out}}\cdot\left(g_{t}\odot\tanh\left(C_{t}\right)\right)\nonumber \\
O_{t} & =\left[D_{t},\left(g_{t}\odot\tanh\left(C_{t}\right)\right)\right],
\end{align}
where $W_{\text{out}}\in\mathbb{R}^{s\times d}$ is a matrix to be
learnt, $\odot$ denotes Hadamard (element-wise) product and $[.,.]$
denotes concatenation. The functions $L_{i}$ are the six NN computations,
given by:

\begin{align}
L^{k}\left(S_{ij}\right) & =S_{ij}T^{k,\left[1:d\right]}S_{ij}^{\top}+I_{0,0}W_{0,0}^{k}S_{ij}^{\top}\nonumber \\
 & \quad+I_{0,1}W_{0,1}^{k}S_{ij}^{\top}+I_{1,0}W_{1,0}^{k}S_{ij}^{\top}\nonumber \\
 & \quad+I_{1,1}W_{1,1}^{k}S_{ij}^{\top}\nonumber \\
S_{ij} & =\left(\left(v_{i},s_{i},e_{i}\right),\left(v_{j},s_{j},e_{j}\right)\right),
\end{align}
where $\left(v_{i},s_{i},e_{i}\right)$ are the $d$ dimensional word
embedding, 6-bit encoding of the syntactic category and an indication
bit of the entity-of-interest for the $i$th phrase, respectively,
$S_{ij}$ encodes the inputs of a left descendant $i$ and a right
descendant $j$ in a parse tree and $k\in\left\{ 1,\ldots,6\right\} $.
Define $D=2d+14$, then $T^{\left[1:d\right]}\in\mathbb{R}^{D\times D\times d}$
\,is a tensor defining bilinear forms, $I_{I,J}$ with $I,J\in\left\{ 0,1\right\} $
are indication functions for having the entity-of-interest on the
left and/or right child and $W_{I,J}\in\mathbb{R}^{d\times D}$ are
matrices to be learnt. 

\begin{figure}
\begin{centering}
\includegraphics[scale=0.5]{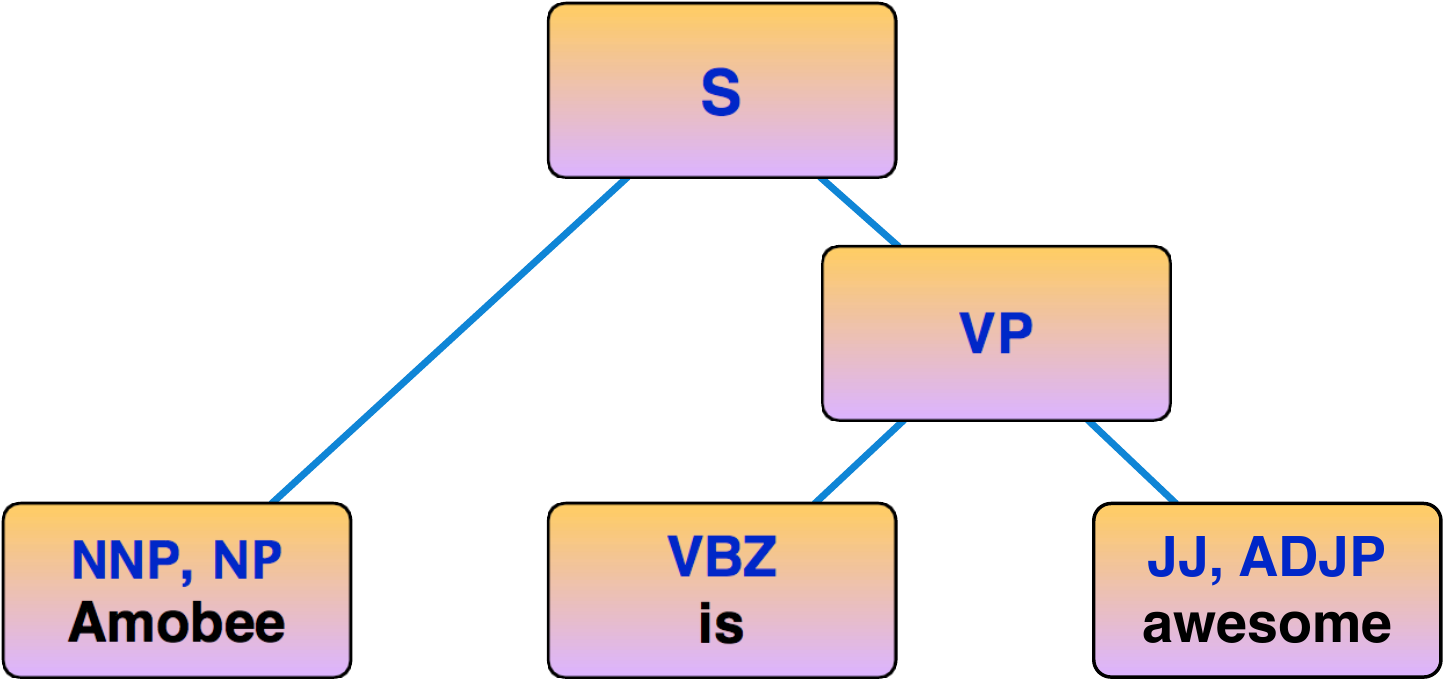}
\par\end{centering}
\caption{\label{fig:amobee-graph}Constituency-based parse tree; the LSTM module
runs on the internal nodes by concatenating the left and right nodes
as its input.}
\end{figure}

The algorithm processes each tweet according to its parse tree, starting
at the leaves and going up combining words into expressions; this
is different than other LSTM algorithms since the parsing data is
used explicitly. As an example, figure \ref{fig:amobee-graph} presents
the simple sentence ``Amobee is awesome'' with its parsing tree.
The leaves are given by $d$-dimensional word vectors together with
their POS tagging, syntactic categories (if defined for the leaf)
and an entity indicator bit. The computation takes place in the inner
nodes; ``is'' and ``awesome'' are combined in a node marked by
``VP'' which is the phrase category. In terms of our terminology,
``is'' and ``awesome'' are the $i,j$ nodes, respectively for
``VP'' node calculation. We define $C''_{t-1}$ as the cell's state
for the \emph{left} child, in this case the ``is'' node. Left and
right are concatenated as input $V_{t}$ and the metadata $D_{t}$
is from the \emph{right} child while $D_{t-1}$ is the metadata from
the \emph{left} child. The second calculation takes place at the root
``S''; the input $V_{t}$ is now a concatenation of ``Amobee''
word vector, the input $O_{t-1}$ holds the $O_{t}$ output of the
previous step in node ``VP''; the cell state $C''_{t-1}$ comes
from the ``Amobee'' node. 

\bibliographystyle{acl_natbib}
\bibliography{semeval2017}

\end{document}